\documentclass[journal]{IEEEtran}

\usepackage{lineno,hyperref}

\usepackage[utf8]{inputenc}
\usepackage[T1]{fontenc}
\usepackage{color}            
\usepackage{subfigure}        
\usepackage{amssymb,amsmath,array}
\usepackage{multirow}

\usepackage[colorinlistoftodos]{todonotes}
\usepackage{multirow}
\usepackage{eurosym}
\usepackage{algorithm}
\usepackage{algorithmic}

\usepackage[figuresright]{rotating}
\modulolinenumbers[1]
\ifCLASSINFOpdf
\else
\fi
\begin{document}
%
\title{Time series clustering based on the\\ characterisation of segment typologies}
%
%
%

\author{David~Guijo-Rubio,
        Antonio Manuel Dur{\'a}n-Rosal,
        Pedro Antonio Guti{\'e}rrez,~\IEEEmembership{Senior Member, IEEE},
        Alicia Troncoso,
        and C{\'e}sar Herv{\'a}s-Mart{\'i}nez,~\IEEEmembership{Senior Member, IEEE}
\thanks{D. Guijo, A. M. Dur{\'a}n-Rosal, P. A. Guti{\'e}rrez, and, C. Herv{\'a}s-Mart{\'i}nez are with the  Department of Computer Science and Numerical Analysis, University of C\'ordoba, Rabanales Campus, Albert Einstein Building 3rd Floor, 14071 C\'ordoba, Spain. E-mails: \{i22gurud, aduran, pagutierrez, chervas\}@uco.es}.
\thanks{A. Troncoso is with the Department of Computer Languages and Systems, University Pablo de Olavide, Sevilla, Spain. E-mail: atrolor@upo.es}
\thanks{ }}

%
%

\markboth{ }%
{Shell \MakeLowercase{\textit{et al.}}: Bare Demo of IEEEtran.cls for IEEE Journals}
%



\maketitle

\begin{abstract}
Time series clustering is the process of grouping time series with respect to their similarity or characteristics. Previous approaches usually combine a specific distance measure for time series and a standard clustering method. However, these approaches do not take the similarity of the different subsequences of each time series into account, which can be used to better compare the time series objects of the dataset. In this paper, we propose a novel technique of time series clustering based on two clustering stages. In a first step, a least squares polynomial segmentation procedure is applied to each time series, which is based on a growing window technique that returns different-length segments. Then, all the segments are projected into same dimensional space, based on the coefficients of the model that approximates the segment and a set of statistical features. After mapping, a first hierarchical clustering phase is applied to all mapped segments, returning groups of segments for each time series. These clusters are used to represent all time series in the same dimensional space, after defining another specific mapping process. In a second and final clustering stage, all the time series objects are grouped. We consider internal clustering quality to automatically adjust the main parameter of the algorithm, which is an error threshold for the segmentation. The results obtained on 84 datasets from the UCR Time Series Classification Archive have been compared against two state-of-the-art methods, showing that the performance of this methodology is very promising.
\end{abstract}


%
\IEEEpeerreviewmaketitle

\label{sec:intro}

\IEEEPARstart{T}{ime} series are an important class of temporal data objects collected chronologically \cite{antunes2001temporal}. Given that they tend to be high dimensional, directly dealing with them in its raw format is very expensive in terms of processing and storage cost, what makes them difficult to analyse. However, time series have applications in many different fields of science, engineering, economics, finance, etc.

In recent years, there has been a high explosion of interest in mining time series databases. Clustering is one of these data mining techniques, where similar data are organized into related or homogeneous groups without specific knowledge of the group definitions \cite{rai2010survey}. 
Usually, clustering is used as a pre-processing step for other data mining tasks.

Time series clustering consists in grouping time series.  There are several recent review papers dealing with time series clustering \cite{Liao2005, Aghabozorgi2015,Fu2011164}. It can be used as a preprocessing step for anomaly detection \cite{leng2009time}, for recognizing dynamic changes in the time series \cite{he2012new}, for prediction \cite{graves2010proximity} and for classification \cite{aghabozorgi2014stock}. For example, the application of these techniques can be used to discover common patterns preceding important paleoclimate events \cite{Nikolaou2014} or mining gene expression patterns \cite{wang2010efficiently}.

Time series clustering can be approached by considering specific distance measures  for time series combined with standard clustering techniques \cite{wang2013experimental,Aghabozorgi2015}. Some of these metrics are designed for equal-length time series, such as the standard Euclidean distance, which is applied to time series in \cite{faloutsos1994fast}, while others, such as the Dynamic Time Warping (DTW) \cite{keogh2002exact, berndt1994using}, can be used for time series of different size. There have been many attempts to obtain better time series distance metrics as extensions of DTW \cite{sakoe1978dynamic,dau2016semi,luczak2016hierarchical,yang2011patterns}. Moreover, apart from adapting distance measures, some authors propose specific adaptations of the clustering algorithm to deal with their special characteristics \cite{yang2011patterns}.

On the other hand, time series segmentation consists in cutting the series in some specific points, trying to achieve two different objectives: (1) dividing time series in segments as a procedure for discovering useful patterns (homogeneous segments) \cite{chung2004evolutionary,tseng2009cluster,Fuchs2009,Nikolaou2014}, or (2) approximating the time series with a set of simple models for each segment without losing too much information \cite{Oliver1997, Oliver1998,Fuchs2010,keogh2003}.

These works of time series segmentation open a new perspective for time series clustering, given that previous time series clustering proposals only search for similarities between the different time series but do not exploit the similarities which can be found in the subsequences of each time series.

In this paper, we propose a novel clustering methodology, which firstly applies time series segmentation via a very fast online polynomial approximation method. Then, unequal-length segments are projected into feature-vectors of equal length, in order to reduce the dimensionality of the original data and have the same length for each mapped segment. A first clustering procedure is applied to group the segments of each time series to recognise similar behaviour segments. Using the results of these clustering processes, and applying a new mapping stage, a second and final clustering process groups the different time series of the dataset. The proposed method is referred to as two-stage statistical segmentation-clustering time series procedure (TS3C). In this way, the method is able to summarise the types of segments that can appear in each individual time series and exploit them to increase the quality of the clustering process.

For adjusting the value of the main parameter of TS3C (which is an error threshold for the segmentation), internal clustering criteria are used, where two different strategies are proposed (considering one single criterion or using a majority voting procedure with a variety of different criteria).

The following advantages can be attributed to TS3C:
\begin{itemize}
	\item It is able to exploit the similarities found in the segments of each individual time series to improve final clustering quality.
	\item It is based on the lowest error approximation of these segments for a particular dataset, allowing the extraction of robust coefficients representing the trend of the segments and their statistical features.
	\item It is domain-independent, not including any special characteristic of the datasets considered.
	\item The formulation is based on two different mapping processes, where the final clustering computational cost does not depend on the size (original number of points) of the time series, but on the number of clusters used to represent it.
	\item The parameter of TS3C is automatically adjusted based on internal criteria.
\end{itemize}

The remainder of the paper is organized as follows. Section \ref{sec:state} summarises the background of time series clustering and the motivation for our clustering approach. Section \ref{sec:clust} describes the algorithm in detail. Section \ref{sec:experiments} presents the experimental results using benchmark time series datasets to show the suitability of the proposed method. Finally, Section \ref{sec:conclusions} concludes the paper and outlines some directions for future research.



\section{Related works} \label{sec:state}

In this section, we begin by reviewing time series clustering methodologies, and the main problems associated to them. Then, we analyse existing clustering evaluation metrics, defining those which are going to be used in our proposal. Finally, time series segmentation methods are also briefly reviewed, given that a time series segmentation method is used as the first step of the methodology proposed.

\subsection{Time series clustering} \label{subsec:stateClust}

There are many works proposed for time series clustering, although their objectives can be very different. Indeed, time series clustering can be classified into three categories \cite{Aghabozorgi2015}:
\begin{itemize}
	\item Whole time series clustering defines each time series as a discrete object and clusters a set of time series measuring their similarity and applying a conventional clustering algorithm.
	\item Subsequence clustering is considered as the clustering of segments obtained from a time series segmentation algorithm. One of its main advantages is that it can discover patterns within each time series.
	\item Time point clustering combine the temporal proximity of time points with the similarity between their corresponding values.
\end{itemize}
We focus on whole time series clustering, which can be applied in three different ways \cite{Aghabozorgi2015}:
\begin{itemize}
	\item Shape-based approach: This method works with the raw time series data, matching, as well as possible, the shapes of the different time series. An appropriate distance measure has to be used, specifically adapted for time series. Then, a conventional clustering algorithm is applied. An example of this approach is that proposed by Paparrizos \textit{et al.} \cite{paparrizos2015k}, which uses a normalized version of the cross-correlation measure (in order to consider the time series shapes) and a method to compute cluster centroids based on the properties of this distance. Policker \textit{et al.} \cite{policker2000nonstationary} presented a model and a set of algorithms for estimating the parameters of a non-stationary time series. This model uses a time varying mixture of stationary sources, similar to hidden markov models (HMMs).Also, Asadi \textit{et al.} \cite{asadi2016creating} proposed a new method based on HMM ensembles, addressing the HMM-based methods problem in separating models of distinct classes.
	\item Feature-based approach: In this case, time series are transformed into a set of statistical characteristics, where the length of this vector is less than the original time series. Each time series is converted into a feature vector of the same length, a standard distance measure is calculated and a clustering algorithm is applied. An example of this approach was presented by R{\"a}s{\"a}nen \textit{et al.} \cite{Rasanen2009}, based on an efficient computational method for statistical feature-based clustering. M{\"o}ller-Levet \textit{et al.} \cite{moller2003fuzzy}, developed a fuzzy clustering algorithm based on the short time series distance (STS), this method being highly sensitive to scale. Hautamaki \textit{et al.} \cite{hautamaki2008time} proposed a raw time series clustering using the dynamic time warping (DTW) distance for hierarchical and partitional clustering algorithms. The problem of DTW is that it can be sensitive to noise.
	\item Model-based approach: Raw time series are converted into a set of model parameters, followed by a model distance measurement and a classic clustering algorithm. McDowell \textit{et al.} \cite{McDowell131151} presented a model-based method, Dirichlet process Gaussian process mixture model (DPGP), which jointly models cluster number with a Dirichlet process and temporal dependencies with Gaussian processes, demonstrating its accuracy on simulated gene expression data sets. Xiong \textit{et al.} \cite{xiong2002mixtures} used a model consisting of mixtures of autoregressive moving average (ARMA) models. This method involves a difficult parameter initialization for the expectation maximization (EM) algorithm. In general, model-based approaches suffer from scalability issues \cite{vlachos2004indexing}. Yang \textit{et al.} \cite{yang2011time} presented an unsupervised ensemble learning approach to time series clustering using a combination of RPCL (rival penalized competitive learning) with other representations.
\end{itemize}



Many of the proposals for time series clustering are based on the combination of a distance measure and a clustering algorithm. First, we will analyse the most important distance measures proposed for time series comparison, and then we will introduce the clustering methods that can be applied based on them\footnote{Further information about time series clustering can be found in \cite{Liao2005} or \cite{Aghabozorgi2015}}.

\subsubsection{Distance measures for time series}  \label{subsec:distances}

Two of the most important distance metrics for time series comparison are the euclidean distance (ED) \cite{faloutsos1994fast} and the dynamic time warping (DTW) \cite{keogh2002exact, berndt1994using}. The first one, ED, compares two time series, $X = \{x_{t}\}_{t = 1}^{N}$ and $Y= \{y_{t}\}_{t = 1}^{N}$, of length $N$ as follows:
\begin{equation}
ED(X, Y) = \sqrt{\sum_{t=1}^{N} (x_t - y_t ) ^2}.
\end{equation}
As can be seen, ED forces both series to have the same length. In contrast, DTW follows the main idea of ED, but applying a local non-linear alignment. This alignment is achieved by deriving a matrix $\mathbf{M}$ with the ED between any two points of $X$ and $Y$. Then, a warping path, $\mathbf{w} = \{w_1, w_2, ..., w_r\}$, is calculated from the matrix of elements $\mathbf{M}$. By using dynamic programming \cite{sakoe1971dynamic}, the previous warping path $\mathbf{w}$ can be computed on matrix $\mathbf{M}$ such as the following condition is satisfied \cite{keogh2002exact}:
\begin{equation}
DTW(X, Y) = \min \sqrt{\sum_{i=1}^{r}w_i}.
\end{equation}

A popular alternative is to constrain the warping path in order to visit only a low number of cells on matrix $\mathbf{M}$ is widely applied \cite{sakoe1978dynamic}.

Recently, Wang \textit{et al.} \cite{wang2013experimental} evaluated $9$ distances measures and demonstrated that DTW is the most accurate distance measure with respect to the rest of measures, while ED is the most efficient one.

Moreover, new distances measures have arisen in recent years. \L uczak \textit{et al.} \cite{luczak2016hierarchical} constructed a new parametric distance function, combining DTW and the derivative DTW distance ($D_{DTW}$) \cite{keogh2001dynamic} (which is computed as the DTW distance considering the derivatives of the time series), where a single real number parameter, $\alpha$,  controls the contribution of each of the two measures to the total value of the combined distances. This distance between time series $X$ and $Y$ is defined as follows:
\begin{equation}
DD_{DTW}(X, Y) = (1 - \alpha)\ DTW(X, Y) + \alpha\ D_{DTW}(X, Y),
\end{equation}
where $\alpha \in [0, 1]$ is a parameter selected by considering the best value for an internal evaluation measure known as inter-group variance ($-V$). This novel metric is shown to outperform the results obtained by $DTW$ and $D_{DTW}$, because it has the advantages of both.

Another state-of-the-art distance measure is based on the invariability to the scale and the translation of the time series and was proposed by Yang \textit{et al.} \cite{yang2011patterns}. This distance between time series $X$ and $Y$ is defined as follows:

\begin{equation}
ISDist(X, Y) = \min_{\alpha, q} \frac{\mid\mid X - \alpha Y_{(q)} \mid\mid}{\mid\mid X \mid\mid},
\end{equation}
where $Y_{(q)}$ is the time series shifted $q$ time units, and $\mid\mid \cdot \mid\mid$ is the $l_2$ norm. $\alpha$ is the scaling coefficient, that could be adjusted to the optimal one by setting the gradient to zero.

\subsubsection{Clustering algorithms}  \label{subsec:clustMethods}

Clustering is a field of data mining based on discovering groups of objects without any form of supervision.

Among the most used metodologies, hierarchical clustering \cite{kaufman2009finding} is based on an agglomerative or a divisive algorithm. The agglomerative approach starts considering each element in a single cluster, and, for each iteration, the pair of clusters with more similarity are merged. On the contrary, the divisive algorithm starts including all elements in a single cluster, and, for each iteration, clusters are divided into smaller subgroups. 

On the other hand, partitional clustering \cite{kaufman2009finding} divides the data into \textit{k} clusters, where each cluster contains at least one element of the dataset. The idea behind this clustering is to minimize the average distance of elements to the cluster centre (also called prototype). Depending on the prototype, there are different algorithms: (1) \textit{k}-means \cite{niennattrakul2007inaccuracies} uses centroids, i.e. the averaged element of the objects does not have to be an object belonging to the dataset, (2) \textit{k}-medoids \cite{hautamaki2008time,vuori2002comparison} uses an object of the cluster as the prototype.

There are also some specific proposals for time series clustering. For example, Wang \textit{et al.} \cite{wang2006characteristic} proposed a method for clustering time series based on their structural characteristics, introducing the following set of features: trend, seasonality, serial correlation, chaos, non-linearity and self-similarity.

\subsection{Clustering evaluation measures} \label{subsec:metrics}

Evaluating the extracted clusters is not a trivial task and has been extensively researched \cite{keogh2003need}. In this paper, we focus on numerical measures, that are applied to judge various aspects of clusters validity \cite{rendon2011internal}.

Different clustering algorithms obtain different clusters and different clustering structures, thus evaluating clustering results is quite important, in order to evaluate clustering structures objectively and quantitatively. There are two different testing criteria \cite{xu2008clustering}: external criteria and internal criteria. External criteria uses class labels (also known as ground truth) for evaluating the assigned labels. Note that the ground truth is not used during the clustering algorithm. On the other hand, internal criteria evaluates the goodness of a clustering structure without respect to external information.

\subsubsection{Internal metrics}

Among the different internal criteria, the most important ones are \cite{Arbelaitz2013}:
\begin{itemize}
	\item Sum of squared error (SSE): This index measures the compactness of a given clustering, independently of the distance to other clusters. ``Better'' clusterings have lower values of SSE. It is defined as:
	\begin{equation}
	SSE = \frac{1}{T}\sum_{i=1}^k \sum_{\mathbf{x} \in \mathcal{C}_i} ED(\mathbf{x},\overline{\mathbf{\mathcal{C}}_i})^2.
	\end{equation} 
	
	\item Normalised sum of squared error (NSSE): This measure look for well-separated groups, maximizing the distance intra-clusters. This can be done by considering the following expression:
	\begin{equation}
	NSSE = \frac{\frac{1}{T}\sum_{i=1}^k \sum_{\mathbf{x} \in \mathcal{C}_i} ED(\mathbf{x},\overline{\mathbf{\mathcal{C}}_i})^2}{(T-1)! \sum_{i=1}^{k}\sum_{j=i+1}^{k} ED(\overline{\mathbf{\mathcal{C}}_i},\overline{\mathbf{\mathcal{C}}_j})}
	\end{equation}
	\item Cali\'nski and Harabasz (CH) \cite{calinski1974dendrite}: This index is defined as the ratio between the internal dispersion of clusters and the dispersion within clusters:
	\begin{equation}
	CH = \frac{\texttt{Tr}(\mathbf{S}_B)\cdot(T-k)}{\texttt{Tr}(\mathbf{S}_W)\cdot (k-1)},
	\end{equation}
	where $T$ is the number of time series and $k$ is the number of cluster used to group segments. Moreover, $\texttt{Tr}(\mathbf{S}_B)$ and $\texttt{Tr}(\mathbf{S}_W)$ are given by:
	\begin{equation}
	\texttt{Tr}(\mathbf{S}_B) = \sum_{i=1}^{k} T_{\mathcal{C}_i} \mid\mid \overline{\mathcal{C}_i} - \overline{Y} \mid\mid ^2,
	\end{equation}
	\begin{equation}
	\texttt{Tr}(\mathbf{S}_W) = \sum_{i=1}^{k} \sum_{y \in \mathcal{C}_i} \mid\mid \overline{y} - \overline{\mathcal{C}_i} \mid\mid ^2,
	\end{equation}
	where, $T_{\mathcal{C}_i}$ is the number of time series that belong to the cluster $\mathcal{C}_i$, $\overline{\mathcal{C}_i}$ is the centroid of cluster $i$, and $\overline{Y}$ is the mean of the time series that belong to the cluster $\mathcal{C}_i$.

	\item Silhouette index (SI) \cite{rousseeuw1987silhouettes}: This measure combines both cohesion and separation, so it is based on the intra-cluster ($a(\mathbf{x}, \mathcal{C}_i)$) and inter-cluster ($b(\mathbf{x}, \mathcal{C}_i)$) distances respectively. This distances are given as follows:
	\begin{equation}
	a(\mathbf{x}, \mathcal{C}_i) = \frac{1}{T_{\mathcal{C}_i}} \sum_{y \in \mathcal{C}_i} ED(\mathbf{x},\mathbf{y}),
	\end{equation}
	\begin{equation}
	b(\mathbf{x}, \mathcal{C}_i) = \min_{\mathcal{C}_l, l \neq i } \left \{ \frac{1}{T_{\mathcal{C}_l}} \sum_{\mathbf{y}\in \mathcal{C}_l} ED(\mathbf{x}, \mathbf{y}) \right \},
	\end{equation}
	where $ED(\mathbf{x}, \mathbf{y})$ is the Euclidean distance between $\mathbf{x}$ and $\mathbf{y}$ time series, as we defined before. Finally, SI index is defined as:
	
	\begin{equation}
	SI = \frac{1}{T} \sum_{i=1}^k \sum_{\mathbf{x} \in \mathcal{C}_i} \frac{ b(\mathbf{x},\mathcal{C}_i) - a(\mathbf{x},\mathcal{C}_i)}{ \max(a(\mathbf{x},\mathcal{C}_i), b(\mathbf{x},\mathcal{C}_i))}.
	\end{equation}

	\item Davies-Bouldin (DB) \cite{davies1979cluster}: The validation of clustering following this measure tries to find compact clusters, which centroids are far away from each other. This index is defined as:
	\begin{equation}
	DB = \frac{1}{k}\sum_{i=1}^k \max_{i\neq j} \frac{\alpha_i + \alpha_j}{ED(\overline{\mathbf{\mathcal{C}}_i},\overline{\mathbf{\mathcal{C}}_j})},
	\end{equation}
	where $\alpha_i$ is the average distance of all elements in cluster $\mathcal{\mathcal{C}}_i$ to centroid $\overline{\mathbf{\mathcal{C}}_i}$, and $ED(\overline{\mathbf{\mathcal{C}}_i},\overline{\mathbf{\mathcal{C}}_j})$ is the euclidean distance between the centroids $\overline{\mathbf{\mathcal{C}}_i}$ and $\overline{\mathbf{\mathcal{C}}_j}$.
	
	\item DUnn index (DU) \cite{dunn1974well}: The Dunn index ponders positively the compact and well-separated clusters. The Dunn index for $k$ clusters $C_i$ with $i=1, ..., k$ is defined as:	
	\begin{eqnarray}
	DU = \min_{i\in\{1,\ldots,l\}} \left(  \min_{j\in\{i+1,\ldots,k\}} \left(  \frac{\delta(\mathcal{C}_i, \mathcal{C}_j)}{M} \right) \right), \\
	M = \max_{m\in\{1,\ldots,k\}} diam(\mathcal{C}_m)
	\end{eqnarray}
	
	where $\delta(\mathcal{C}_i, \mathcal{C}_j)$ is the dissimilarity between clusters $\mathcal{C}_i$ and $\mathcal{C}_j$, and $diam(\mathcal{C}_m)$ is the diameter of the cluster $\mathcal{C}_m$, which are given as follows:
	\begin{eqnarray}
	\delta(\mathcal{C}_i, \mathcal{C}_j) = \min_{\mathbf{x} \in \mathcal{C}_i,\;\; \mathbf{y} \in \mathcal{C}_j} \mid\mid \mathbf{x}-\mathbf{y} \mid\mid,\\
	diam(\mathcal{C}_m) = \max_{\mathbf{x}, \mathbf{y} \in \mathcal{C}_m} \mid\mid \mathbf{x}-\mathbf{y} \mid\mid.
	\end{eqnarray}
	
	The Dunn index is very sensitive to noise, and different variants have been considered. We chose the three variants that had betters results in \cite{Arbelaitz2013}, where are referred to as GD33, GD43 and GD53. These variants have the following equations for $\delta(\mathcal{C}_i, \mathcal{C}_j)$, respectively:
	\begin{align}
	\delta(\mathcal{C}_i, \mathcal{C}_j) = \frac{1}{N_{\mathcal{C}_i}\ N_{\mathcal{C}_j}} \sum_{\mathbf{x} \in \mathcal{C}_i} \sum_{\mathbf{y} \in \mathcal{C}_j} ED(\mathbf{x}, \mathbf{y}),\\
	\delta(\mathcal{C}_i, \mathcal{C}_j) = ED(\overline{\mathbf{\mathcal{C}}_i}, \overline{\mathbf{\mathcal{C}}_j}),
	\end{align}
	\vspace{-0.5cm}
	\begin{align}
	\delta(\mathcal{C}_i, \mathcal{C}_j) = \frac{1}{N_{\mathcal{C}_i}+ N_{\mathcal{C}_j}} \cdot & \left( \sum_{\mathbf{x} \in \mathcal{C}_i} ED(\mathbf{x}, \overline{\mathbf{\mathcal{C}}_i}) +\right.\\
	 &\left. +\sum_{\mathbf{y} \in \mathcal{C}_j} ED(\mathbf{y}, \overline{\mathbf{\mathcal{C}}_j}) \right).\nonumber
	\end{align}
	
	For the last variant (GD53), a new definition of $diam(\mathcal{C}_m)$ is included:
	\begin{equation}
	diam(\mathcal{C}_m) = \frac{2}{N_{\mathcal{C}_i}} \sum_{\mathbf{x} \in \mathcal{C}_i} d^*_{ps} (\mathbf{x}, \mathcal{C}_i),
	\end{equation}
	where $d^*_{ps} (\mathbf{x}, \mathcal{C}_i)$ is the Point Symmetry-Distance between the object $\mathbf{x}$ and the cluster $\mathcal{C}_i$\footnote{For further information, see \cite{Arbelaitz2013}}.
	
	\item COP index (COP): This index uses the distance from the points to their cluster centroids and the furthest neighbour distance. The equation is the following:
	\begin{equation}
	COP = \frac{1}{T} \sum_{i=1}^k \frac{\sum_{\mathbf{y} \in \mathcal{C}_i}\ ED(\mathbf{y}, \overline{\mathbf{\mathcal{C}}_i})}{N_{C_i} \cdot \min_{\mathbf{x} \notin C_i}\ max_{\mathbf{y} \in C_i}\ ED(\mathbf{x}, \mathbf{y})}
	\end{equation} 
	
\end{itemize}

CH, SI, COP, DU and variants have to be maximised. Conversely, DB, SSE and NSSE have to be minimised. The most common measures in the literature are CH, DU and SSE. The work of Arbelaitz \textit{et al.} \cite{Arbelaitz2013} compares 30 cluster validity indices in many different environments and demonstrated that CH and DU behave better than the other indices.

\subsubsection{External metrics}

On the other hand, external indices measure the similarity between the cluster assignment and the ground truth, which has to be given as a form of evaluation but should not be used during the clustering. There are many metrics in the literature \cite{manning1995introduction}, although the most widely used is the rand index (RI) \cite{rand1971objective}. This measure penalizes false positive and false negative decisions during clustering. RI is given as:
\begin{equation}
RI = \frac{a+b}{a+b+c+d},
\end{equation}
where $a$ is the number of time series that are assigned to the same cluster and belong to the same class (according to ground truth), $b$ is the number of time series that are assigned to different clusters and belong to different classes, $c$ is the number of time series that are assigned to different clusters but belong to the same class, and $d$ is the number of time series that are assigned to the same cluster but belong to different classes.

\subsection{Time series segmentation} \label{subsec:stateSegm}

One of the steps of our proposal is based on dividing each time series in a sequence of segments. This is known as time series segmentation, which consists in cutting the time series in some specific points, trying to achieve different objectives, where, as mentioned before, the two main points of view are:
\begin{itemize}
	\item Discovering similar patterns: The main objective is the discovery and characterization of important events in the time series, by obtaining similar segments. The methods of Chung \textit{et al.} \cite{chung2004evolutionary}, Tseng \textit{et al.} \cite{tseng2009cluster} and  Nikolaou \textit{et al.} \cite{Nikolaou2014} are all based on evolutionary algorithms, given the large size of the search space when deciding the cut points.
	
	\item Approximating the time series by a set of simple models, e.g. linear interpolation or polynomial regression: These methods could also be considered as representation methods. The main goal of these methods is to summarize a single time series, in order to reduce the difficulty of processing, analysing or exploring large time series, approximating the segments obtained by linear models. Keogh \textit{et al.} \cite{keogh2003} proposed some methods which use linear interpolations between the cut points. Oliver \textit{et al.} \cite{Oliver1997,Oliver1998} develop a method that detect points with high variation and, then, replace each segment with the corresponding approximation.  Finally, the method proposed by Fuchs \textit{et al.} \cite{Fuchs2010} is a growing window procedure (known as SwiftSeg), which returns unequal-length segments based on a online method. SwiftSeg is very fast, simultaneously obtaining a segmentation of the time series and the coefficients of the polynomial least squares approximation, the computational cost depending only on the degree of the polynomial instead of the window length. When compared to many other segmentation methods, SwiftSeg is shown to be very accurate while involving a low computational cost \cite{Fuchs2010}.
\end{itemize}


\section{A two-stage statistical segmentation-clustering time series procedure (TS3C)} \label{sec:clust}

Given a time series clustering dataset, $D = \{Y_i\}_{i=1}^{T}$, where $Y_i = \{y_t\}_{t = 1}^{N_i}$ is a time series of length $N_i$, the objective of the proposed algorithm is to organize them into $L$ groups,  $\mathcal{G} = \left\lbrace \mathcal{G}_1, \mathcal{G}_2, \ldots \mathcal{G}_L \right\rbrace$, optimizing the clustering quality, where $\forall \mathcal{G}_i\neq \mathcal{G}_j, \mathcal{G}_i\cap\mathcal{G}_j=\emptyset$ and $\bigcup_{l=1}^{L} \mathcal{G}_{l} = \mathcal{G}$.

The algorithm is based on two well-identified stages. The first stage is applied individually to each time series and acts as a dimensionality reduction. It consists of a segmentation procedure and a clustering of segments, discovering common patterns of each time series. The second clustering stage is applied to the mapped time series to discover the groups. The main steps of the algorithm are summarized in \figurename{ \ref{fig:pseudocode}}.

\begin{figure}[b]
	\centering
	\begin{algorithmic}[1]
		\item[] \hspace{-0.7cm}\underline{\textbf{Time series clustering}}:
		\REQUIRE Time series dataset
		\ENSURE Best quality clustering
		\FOR{Each time series}
		\STATE Apply time series segmentation
		\FOR{Each segment}
		\STATE Extract the coefficients of the segment
		\STATE Compute the statistical features
		\STATE Combine the coefficients and the statistical features into a single array
		\ENDFOR
		\STATE Cluster all the mapped segments
		\STATE Based on the previous clustering, map each time series
		\ENDFOR
		\STATE Cluster mapped time series
		\STATE Evaluate the goodness of the clustering
		\RETURN Best quality clustering
	\end{algorithmic}
	\caption{Main steps of the TS3C algorithm.}
	\label{fig:pseudocode}
\end{figure}

\subsection{First stage}

\begin{figure}[htp]
	\centering
	\includegraphics[keepaspectratio,width=9.5cm]{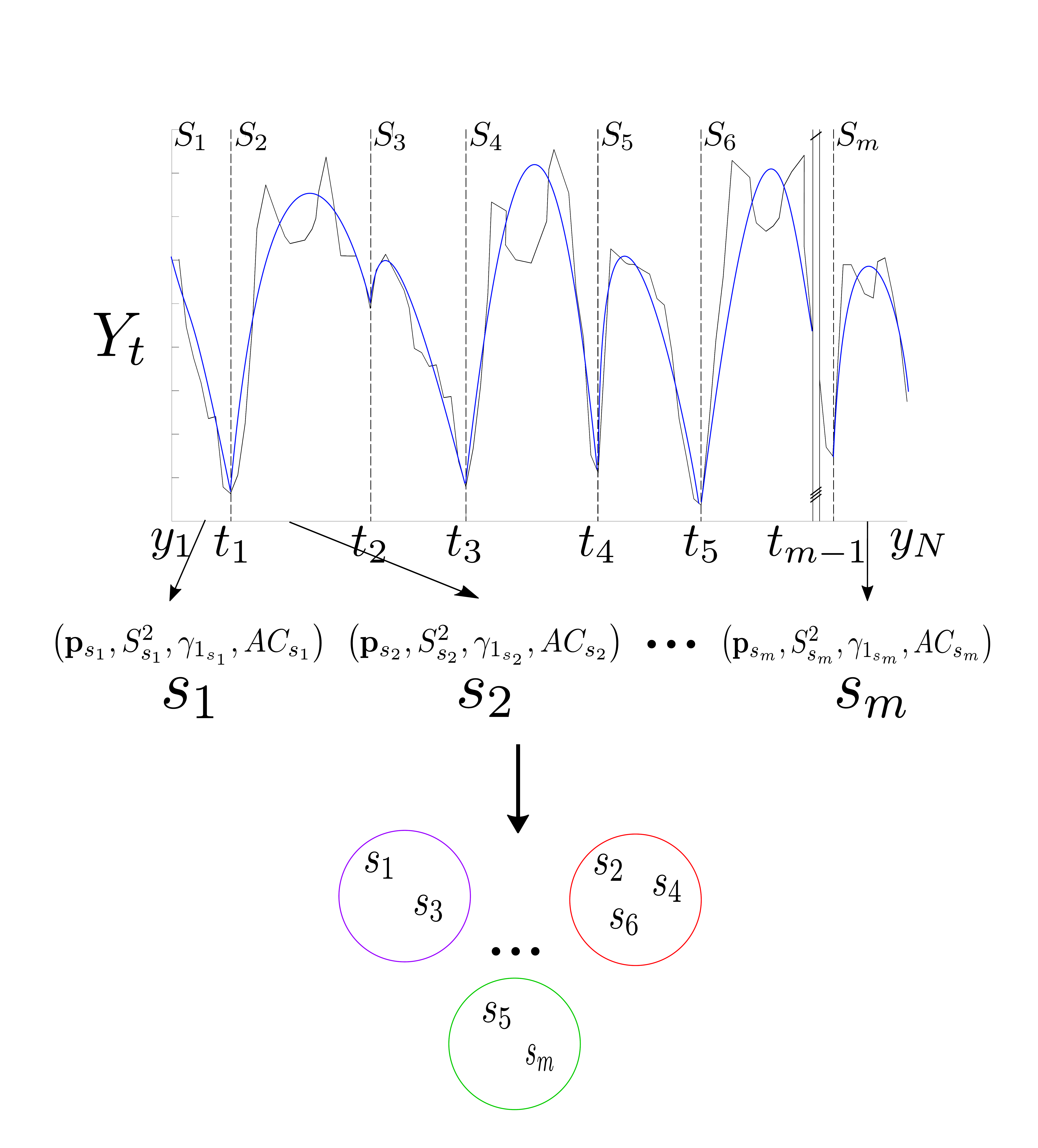}
	\caption{The first stage consists of three steps, applied to each time series of the database $D$: firstly, a segmentation procedure is applied to the time series $Y_t$. Then, segments extracted are mapped into a $(c+f)$-dimensional space. Finally, these arrays are cluster into $k$ groups.}
	\label{fig:firststage}
\end{figure}

The first stage of TS3C consists of a time series segmentation, the extraction of statistical features of each segment, and the clustering of the segments for each time series. The steps of the first stage  can be checked in Figure \ref{fig:firststage}.

\subsubsection{Time series segmentation}\label{subsub:segmentation}

In general, segmentation problems are used for discovering cut points in the time series to achieve different objectives. For a given time series of length $N_i$, the segmentation consists in finding $m$ segments defined by $\mathbf{t}=\{t_s\}_{s=1}^{m-1}$ cut points. In this way, the set of segments $\mathbf{S}= \{s_{1}, s_{2}, \ldots, s_{m}\}$ are formed by: $s_{1} = \{y_1, \ldots, y_{t_1}\}, s_{2} = \{y_{t_1}, \ldots, y_{t_2}\}, \ldots, s_{m}=\{y_{t_{m-1}}, \ldots, y_{N_i}\}$. Specifically, in this paper, we apply SwiftSeg, a growing window procedure proposed in \cite{Fuchs2010}. The algorithm iteratively introduces points of the time series into a growing window and simultaneously updates the corresponding least-squares polynomial approximation of the segment and its error. The window grows until an error threshold is exceeded. When this happens, a cut point ($t_s$) is included and the segment is finished. The process is repeated until reaching the end of the time series. We consider the following error function (standard error of prediction, $SEP$):
\begin{equation}
SEP_s = \frac{\sqrt{SSE_s}}{|\overline{Y_s}|},
\end{equation}where, $SSE_s$ stands for Sum of Squared Errors of segment $s$, and $|\overline{Y_s}|$ is the average value of segment $s$. $SSE_s$ and $\overline{Y_s}$ are defined as:
\begin{equation}
SSE_s = \sum_{i = t_{s-1}}^{t_s} (\hat{y_{i}} - y_{i})^2,
\end{equation}
\begin{equation}
\overline{Y_s} = \frac{1}{t_s - t_{s-1} + 1} \sum_{i = t_{s-1}}^{t_s} y_i,
\end{equation}
where, $y_i$ is the time series value at time $i$, and $\hat{y_{i}}$ is its corresponding least-squares polynomial approximation. The advantage of this error function is that it does not take into account the scale of the values of each segment. The maximum error from which the window is not further grown is denoted as $SEP_\mathrm{max}$ and has to be defined by the user.


\subsubsection{Segment mapping} \label{subsub:static}

After the segmentation process, each segment is mapped to an array, including the polynomial coefficients of the least squares approximation of the segment and a set of statistical features. Thus, each segment is projected into a $l$-dimensional space, where $l$ is the length of the mapped segment.

The coefficients are directly obtained from the update procedure of the time series segmentation growing window specified in \cite{Fuchs2010}. We discard the intercept, given that we are interested in shape of the segment, not in its relative value.

Moreover, we compute the following statistical features:
\begin{enumerate}
	\item The variance ($S_s^2$) measures the variability of the segment:
	\begin{equation}
	\textstyle S_s^2 = \frac{1}{t_s-t_{s-1}+1} \sum_{i=t_{s-1}}^{t_s} \left({y_{i}} - \overline{y_{s}}\right)^2,
	\end{equation} where ${y_{i}}$ are the time series values of the segment, and $\overline{y_{s}}$ is the average of the values of the segment $s$.
	
	\item The skewness ($\gamma_{1s}$) represents the asymmetry of the distribution of the time series values in the segment with respect to the arithmetic mean:
	\begin{equation}
	\textstyle\gamma_{1s} = \frac{\frac{1}{t_s-t_{s-1}+1}\sum _{i={t_{s-1}}}^{t_s} (y_i - \overline{y_s})^3}{\hat{\sigma}_s^3},
	\end{equation}where $\hat{\sigma}_s$ is the standard deviation of the $s$-th segment. 
	
	\item The autocorrelation coefficient ($AC_s$) is a measure of the correlation between the current values of the time series and the previous ones:
	\begin{equation}
	\textstyle AC_s =  \frac{\sum_{i=t_{s-1}}^{t_s} (y_i - \overline{y_s}) \cdot (y_{i+1} - \overline{y_s})} {{S_{s}^2}}.
	\end{equation}
\end{enumerate}

Using these statistical features and the coefficients previously extracted, each segment is mapped into a $l$-dimensional array ($l = c+f$), which is used as the segment representation, where $c$ is the degree of the polynomial and $f$ is the number of statistical features ($f=3$, in our case). The mapping is then defined by:
\begin{equation}
\mathbf{v}_s = (\mathbf{p}_s, S^2_s, \gamma_{1_s}, AC_s),
\end{equation}
where $\mathbf{p}_s$ are the parameters of the polynomial approximation that approximates the segment $s$. This procedure is able to reduce the length of the segment from $(t_s - t_{s-1} + 1)$ to $\left( c+f\right) $.

\subsubsection{Segment clustering}\label{sec:segmentClustering}

A hierarchical clustering is subsequently applied to group all the segments of a time series, represented by the set of arrays $\left\{\mathbf{v}_s, s\in\{1,\ldots,m\}\right\}$. The main goal is representing all the time series with arrays of the same length and significantly reducing the size of the representation.

The hierarchical clustering used is agglomerative, using the Ward distance defined in \cite{ward1963hierarchical} as the similarity measure. The number of clusters considered for segment mapping is $k=2$, for all the datasets and time series. This value is found to be robust enough for extracting a minimum amount of information about the internal characteristics of the series.


\subsection{Second stage}

\begin{figure}[htp]
	\centering
	\includegraphics[keepaspectratio,width=9cm]{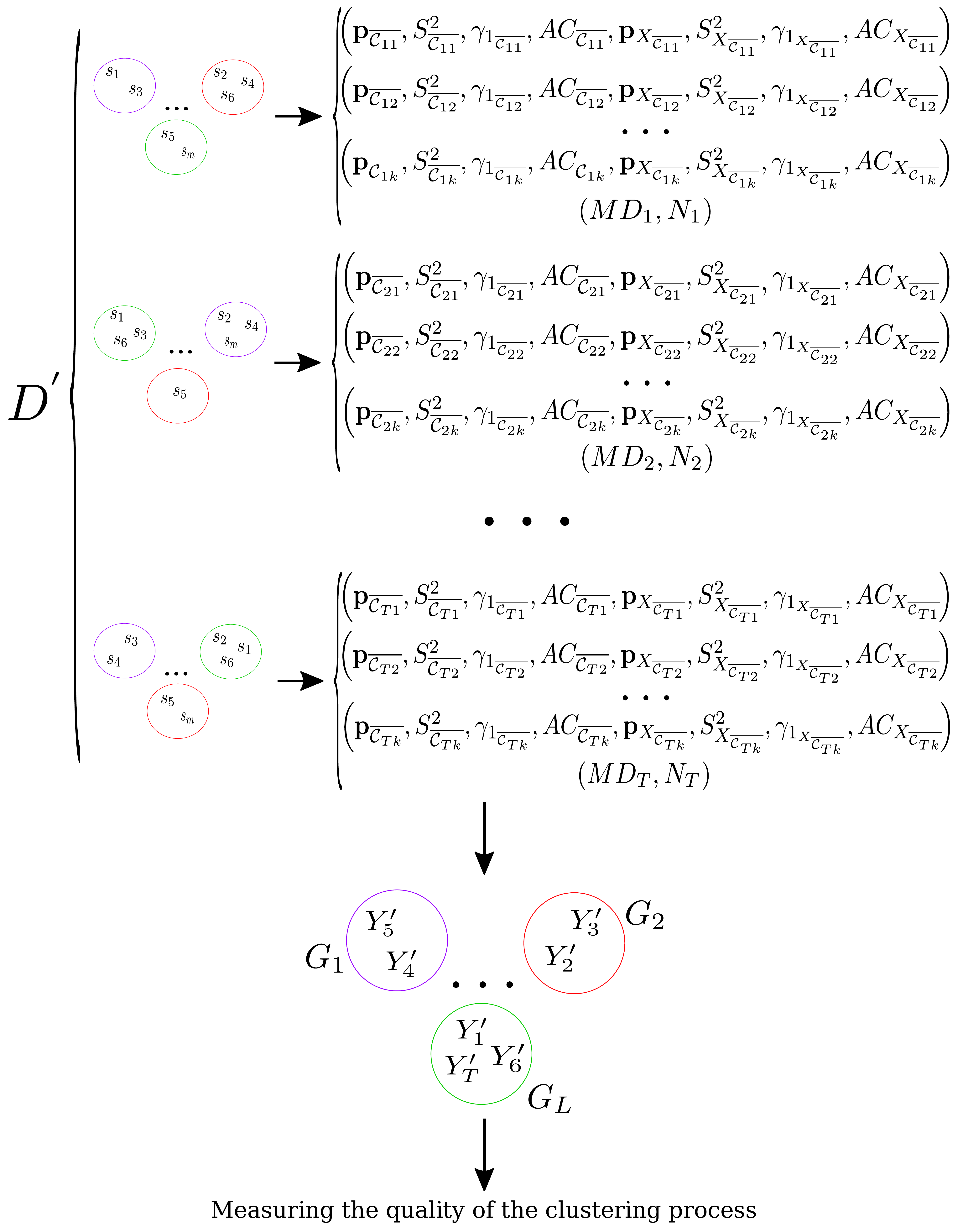}
	\caption{The second stage consists of four steps: firstly, each cluster is represented by a set of statistical features, which, in conjunction, represents the mapped time series, $Y'_t$. Then, a clustering process is applied to mapped time series, clusters being denoted as $G_l$. After that, the measurement of the clustering quality is performed, using different strategies based on internal indices to choose the best configuration of $SEP_\mathrm{max}$. Finally, an external index compares our approach to the ground truth.}
	\label{fig:secondstage}
\end{figure}

The second stage of the method proposed consists of mapping the time series to a common representation, clustering them and evaluating its quality. The steps of the second stage are summarised in Figure \ref{fig:secondstage}.

\subsubsection{Time series mapping}

The first stage transform each time series to a set of clustered segments. Now, a specific mapping process is used to represent all time series in the same dimensional space.

For each time series, $Y_i$, we extract the corresponding centroids from the process described in Section \ref{sec:segmentClustering}, $\overline{\mathcal{C}_{ij}}$, where $i\in\{1,\ldots,T\}$, $j\in\{1,\ldots,k\}$, $k$ being the number of clusters and $T$ being the number of time series. For each cluster, $\mathcal{C}_{ij}$, we extract:
\begin{itemize}
	\item Its centroid $\overline{\mathcal{C}_{ij}}$, i.e. the average of all cluster points.
	\item The mapping of the segment with higher variance, denoted as $X_{\mathcal{C}_{ij}}$ (in order to represent the extreme segments, i.e. the most characteristic segment of the cluster $\mathcal{C}_{ij}$).
\end{itemize}
In this way, the length of the mapped cluster is $w = (l \times 2)$, where, $(l \times 2)$ is the length of both the centroid and the extreme segment. This process is applied to each cluster of each time series. The mapping process of a centroid can be formally specified as:
\begin{equation}
(\overline{\mathcal{C}_{ij}}, X_{\mathcal{C}_{ij}}), \forall i\in\{1,\ldots,T\}, \forall j\in\{1,\ldots,k\}.
\end{equation}

Apart from the representation of each cluster, two more characteristics of the time series are also considered:
\begin{itemize}
	\item The error difference ($MD_{\mathcal{C}_{i}}$) between the segment least similar to its centroid (farthest segment) and the segment most similar to its centroid (closest segment). We evaluate the error of a segment by using the Mean Squared Error (MSE) of the corresponding polynomial approximation.
	\item The number of segments of the time series, $N_{\mathcal{C}_{i}}$.
\end{itemize}

The order in which the clusters are arranged in the mapping is important and has to be consistent along all the time series. This is done by a simple matching procedure, where the centroids of one time series are used as reference, and, for the rest of time series, the closest centroids with respect to the reference ones are matched together.

Once the matching is defined, each time series is transformed into a mapped time series $Y'_i$, composed by the characteristics of the extracted clusters. Thus, the length of a mapped time series is $\left( w \times k\right)  + v$, $k$ being the number of clusters, and $v$ being the number of the extra information for the time series, which is 2 in our case.

\subsubsection{Time series clustering}

In this step, the algorithm receives the mapped time series and the clustering is performed, choosing again an agglomerative hierarchical methodology. The idea is to group similar time series in the same cluster. In our experiments, the number of clusters to be found, $L$, is defined as the number of classes of the dataset (given that we consider time series classification datasets for the evaluation). In a real scenario, $L$ should be given by the user. This advantage is given to all the methods compared.

\subsection{Parameter adjustment}\label{subsec:parametersAdjustment}

The TS3C algorithm previously defined involves only one important parameter that has to be adjusted by the user: the error threshold for the segmentation procedure, $SEP_\mathrm{max}$ (see Section \ref{subsub:segmentation}). We propose to adjust it considering internal clustering evaluation metrics (see Section \ref{subsec:metrics}), which can be used without knowing the ground truth labels.

In this way, the algorithm is run using a set of values for this parameter, all these cases being evaluated in terms of these internal measures. Two different strategies are proposed to select the best parameter value:
\begin{itemize}
	\item Selecting the $SEP_\mathrm{max}$ leading to the best Cali\'nski and Harabasz index (CH), given that this index has been proved to be very robust \cite{Arbelaitz2013}.
	\item Selecting the $SEP_\mathrm{max}$ which obtains the best value for the highest number of internal measures. All the internal metrics defined in Section \ref{subsec:metrics} are used in this case. We refer to this option as majority voting.
\end{itemize}

\section{Experimental results and discussion}
\label{sec:experiments}
In this section, the experiment results are presented and discussed. Firstly, we detail the characteristics of the datasets used in the experiments. Secondly, we explain the experimental setting. Then, we show the results and discuss them. Finally, an statistical analysis of the results is performed.

\subsection{Datasets}\label{subsection:datasets}

84 datasets from the UCR Time Series Classification Archive \cite{UCRArchive} have been considered. This benchmark repository (last updated in Summer 2015) is made of synthetic and real datasets of different domains. The repository was originally proposed for time series classification, so that each dataset was split into training and test subsets. However, for time series clustering, where the class label will only be considered for evaluating the clustering quality, we can safely merge these subsets. The details of the datasets are included in Table \ref{table:datasets}. Also, we have computed the Imbalance Ratio (IR) for each dataset, as the ratio of the number of instances in the majority class with respect to the number of examples in the minority class \cite{garcia2012effectiveness}. Although the length of the time series is the same for all elements of each dataset of the repository, the TS3C algorithm could be applied to datasets with different-length time series.

\begin{table*}[htp]
	\centering
	\caption{Characteristics of the datasets used in the experiments. \#CL: number of classes, \#EL: number of elements (size), LEN: time series length, \%IR: imbalanced ratio.}
	\label{table:datasets}
	\resizebox{1\textwidth}{!}{
		\begin{tabular}{lcccc||lcccc}
			\hline
			\hline
			Dataset                      & \#CL & \#EL  & LEN & \%IR & Dataset & \#CL & \#EL & LEN & \%IR \\ \hline
			50words (50W)                      & 50   & 905   & 270 & $18.167$ & MedicalImages  (MED)                & 10   & 1141 & 99  & $25.826$ \\
			Adiac (ADI)                       & 37   & 781   & 176 & $1.450$ & MiddlePhalanxOutlineAgeGroup (MPA)  & 3    & 554  & 80  & $2.891$ \\
			ArrowHead (ARR)                   & 3    & 211   & 251 & $1.246$ & MiddlePhalanxOutlineCorrect (MPC)   & 2    & 891  & 80  & $1.644$ \\
			Beef         (BEE)                & 5    & 60    & 470 & $1.000$ & MiddlePhalanxTW  (MPT)              & 6    & 553  & 80  & $5.941$ \\
			BeetleFly  (BFL)                  & 2    & 40    & 512 & $1.000$ & MoteStrain  (MOT)                   & 2    & 1272 & 84  & $1.167$ \\
			BirdChicken  (BIR)                & 2    & 40    & 512 & $1.000$ & NonInvasiveFatalECG\_Thorax1 (NO1)  & 42   & 3765 & 750 & $1.307$ \\
			Car   (CAR)                       & 4    & 120   & 577 & $1.000$ & NonInvasiveFatalECG\_Thorax2 (NO2)   & 42   & 3765 & 750 & $1.307$ \\
			CBF  (CBF)                        & 3    & 930   & 128 & $1.000$ & OliveOil  (OLI)                     & 4    & 60   & 570 & $3.125$ \\
			ChlorineConcentration  (CHL)      & 3    & 4307  & 166 & $2.203$ & OSULeaf  (OSU)                      & 6    & 442  & 427 & $2.553$ \\
			CinC\_ECG\_torso (CIN)            & 4    & 1420  & 1639 & $1.000$ & PhalangesOutlinesCorrect (PHA)      & 2    & 2658 & 80 & $2.115$ \\
			Coffee  (COF)                     & 2    & 56    & 286 & $1.075$ & Phoneme  (PH0)                      & 39   & 2110 & 1024 & $119.000$ \\
			Computers   (COM)                 & 2    & 500   & 720 & $1.000$ & Plane (PLA)                          & 7    & 210  & 144 & $1.000$ \\
			Cricket\_X (CRX)                  & 12   & 780   & 300 & $1.000$ & ProximalPhalanxOutlineAgeGroup (PPA) & 3    & 605  & 80  & $3.247$ \\
			Cricket\_Y (CRY)                   & 12   & 780   & 300 & $1.000$ & ProximalPhalanxOutlineCorrect (PPC) & 2    & 891  & 80  & $2.115$ \\
			Cricket\_Z (CRZ)                  & 12   & 780   & 300 & $1.000$ & ProximalPhalanxTW  (PPT)            & 6    & 605  & 80  & $14.000$ \\
			DiatomSizeReduction (DIA)         & 4    & 322   & 345 & $2.912$ & RefrigerationDevices (REF)          & 3    & 750  & 720 & $1.000$ \\
			DistalPhalanxOutlineAgeGroup  (DPA )& 3    & 539   & 80  & $7.126$ & ScreenType  (SCR)                   & 3    & 750  & 720 & $1.000$ \\
			DistalPhalanxOutlineCorrect (DPC) & 2    & 876   & 80  & $1.600$ & ShapeletSim (SHS)                   & 2    & 200  & 500 & $1.000$ \\
			DistalPhalanxTW (DPT)             & 6    & 539   & 80  & $9.808$ & ShapesAll (SHA)                     & 60   & 1200 & 512 & $1.000$ \\
			Earthquakes  (EAR)                & 2    & 461   & 512 & $3.957$ & SmallKitchenAppliances (SMA)        & 3    & 750  & 720 & $1.000$ \\
			ECG200   (EC2)                    & 2    & 200   & 96  & $1.985$ & SonyAIBORobotSurface (SO1)          & 2    & 621  & 70  & $1.283$ \\
			ECG5000 (EC5)                     & 5    & 5000  & 140 & $121.625$ & SonyAIBORobotSurfaceII (SO2)         & 2    & 980  & 65  & $1.606$ \\
			ECGFiveDays  (ECF)                & 2    & 884   & 136 & $1.000$ & StarLightCurves (STA)               & 3    & 9236 & 1024 & $4.008$ \\
			ElectricDevices  (ELE)            & 7    & 16637 & 96  & $3.415$ & Strawberry (STR)                    & 2    & 983  & 235 & $1.800$ \\
			FaceAll (FAA)                     & 14   & 2250  & 131 & $6.813$ & SwedishLeaf (SWE)                   & 15   & 1125 & 128 & $1.000$ \\
			FaceFour (FAF)                    & 4    & 112   & 350 & $1.545$ & Symbols      (SYM)                  & 6    & 1020 & 398 & $1.117$ \\
			FISH  (FIS)                       & 7    & 350   & 463 & $1.000$ & synthetic\_control (SYN)             & 6    & 600  & 60  & $1.000$ \\
			FordA  (FOA)                      & 2    & 4921  & 500 & $1.056$ & ToeSegmentation1 (TO1)               & 2    & 268  & 277 & $1.094$ \\
			FordB  (FOB)                      & 2    & 4446  & 500 & $1.035$ & ToeSegmentation2  (TO2)             & 2    & 166  & 343 & $2.952$ \\
			Gun\_Point (GUN)                  & 2    & 200   & 150 & $1.000$ & Trace  (TRA)                        & 4    & 200  & 275 & $1.000$ \\
			Ham  (HAM)                        & 2    & 214   & 431 & $1.078$ & Two\_Patterns  (TWP)                & 4    & 5000 & 128 & $1.087$ \\
			HandOutlines  (HAN)               & 2    & 1370  & 2709 & $1.768$ & TwoLeadECG  (TWE)                   & 2    & 1162 & 82 &  $1.000$ \\
			Haptics (HAP)                     & 5    & 463   & 1092 & $1.282$ & uWaveGestureLibrary\_X  (UWX)       & 8    & 4478 & 315 & $1.002$ \\
			Herring (HER)                     & 2    & 128   & 512 & $1.510$ & uWaveGestureLibrary\_Y (UWY)        & 8    & 4478 & 315 & $1.002$ \\
			InlineSkate  (INL)                & 7    & 650   & 1882 & $1.887$ & uWaveGestureLibrary\_Z  (UWZ)       & 8    & 4478 & 315 & $1.002$ \\
			InsectWingbeatSound (INS)         & 11   & 2200  & 256 & $1.000$ & UWaveGestureLibraryAll  (UWA)       & 8    & 4478 & 945 & $1.002$ \\
			ItalyPowerDemand (ITA)            & 2    & 1096  & 24  & $1.003$ & wafer          (WAF)                & 2    & 7164 & 152 & $8.402$ \\
			LargeKitchenAppliances (LAR)      & 3    & 750   & 720 & $1.000$ & Wine           (WIN)                & 2    & 111  & 234 & $1.056$ \\
			Lighting2 (LI2)                   & 2    & 121   & 637 & $1.521$ & WordsSynonyms   (WOS)               & 25   & 905  & 270 & $16.667$ \\
			Lighting7  (LI7)                  & 7    & 143   & 319 & $2.714$ & Worms  (WOR)                        & 5    & 258  & 900 & $4.360$ \\
			MALLAT     (MAL)                  & 8    & 2400  & 1024 & $1.000$ & WormsTwoClass (WOT)                  & 2    & 258  & 900 & $1.367$ \\
			Meat           (MEA)              & 3    & 120   & 448 & $1.000$ & yoga   (YOG)                        & 2    & 3300 & 426 & $1.157$\\
			\hline
			\hline
		\end{tabular}
	}
\end{table*}



\subsection{Experimental setting}\label{sec:experimental}
The experimental design for the datasets under study is presented in this subsection.

The degree of the polynomial of the least-square approximation is set to $1$, given that higher order polynomials led to worse results. The number of groups for the segment clustering is $k=2$, given that the nature of the different time series datasets seems to be very similar. The other parameter of the algorithm, $SEP_\mathrm{max}$ has been adjusted using the two options described in Section \ref{subsec:parametersAdjustment}: (1) directly selecting the clustering leading to the best Cali\'nski and Harabasz (CH) measure (TS3C$_{CH}$), and (2) considering all the internal measures in Section \ref{subsec:metrics} and applying a majority voting procedure to select the best one (TS3C$_{MV}$). The range considered for the parameter $SEP_\mathrm{max}$ is the following one $\{10,20,30,\ldots,100\}$.

The Rand Index (RI) is used as external measure for evaluating the results. The number of clusters (for the time series clustering stage) is set to the number of real labels in each dataset. 

We compare our method against two state-of-the-art algorithms:
\begin{itemize}
	\item $DD_{DTW}$ distance metric together with a hierarchical clustering algorithm ($DD_{DTW}$-HC) \cite{luczak2016hierarchical}. This method considers the negative intergroup variance ($-V$) as the internal cluster validation measure to set the $\alpha$ value (see Section \ref{subsec:distances}). This is the best technique from those proposed in \cite{luczak2016hierarchical}.
	\item \textit{K}-Spectral Centroid (KSC). This algorithm, proposed by Yang \textit{et al.} \cite{yang2011patterns}, is able to find clusters of time series that share a distinct temporal pattern. See more details in Sections \ref{subsec:distances} and \ref{subsec:clustMethods}. 
\end{itemize}

Because KSC algorithm is stochastic, it was run $30$ times, while the rest of methods (TS3C$_{CH}$,  TS3C$_{MV}$ and $DD_{DTW}$-HC) are deterministic (and they have been run once). The computational time needed by all the algorithms will also be analysed in this section.

\subsection{Results} \label{subsection:discussion}

The results of TS3C$_{CH}$ and TS3C$_{MV}$ are shown in Table \ref{table:resultsMV}, including both the RI performance the computational time needed for the algorithms (average computational time in case of KSC). Note that for some datasets, the running time of $DD_{DTW}$-HC was higher than $763587$ (maximum time of the rest of methods), so that they have been marked with ``$>763587$'' and the results have been taken from \cite{luczak2016hierarchical}. As can be seen, we have included, as a subscript, the error threshold for the segmentation algorithm ($SEP_{\max}$) of the best clustering configuration for the TS3C$_{CH}$ and TS3C$_{MV}$ methods (obtained using internal criteria).

From the results in Table \ref{table:resultsMV}, the following facts can be highlighted:
\begin{itemize}
	\item Compared with $DD_{DTW}$-HC, TS3C$_{CH}$ obtains better solutions for $48$ datasets, slightly worse results for $34$, and obtains the same solution for the remaining $2$ datasets. If $DD_{DTW}$-HC is compared with TS3C$_{MV}$, our approach obtains better solutions in $50$ datasets, worse results for $32$, and similar results for the remaining $2$ datasets.
	\item Compared with KSC,  TS3C$_{CH}$ leads to better solutions in $42$ datasets, while in $41$ the results are slightly worse. Finally, for the remaining dataset, the result is the same. When this method is compared with TS3C$_{MV}$, better solutions are obtained in $45$ cases, slightly worse solutions are found for $37$ datasets, and no differences for $2$ datasets.
\end{itemize}
Analysing average performance, the mean RI values are $0.661$, $0.657$, $0.606$ and $0.601$, for TS3C$_{CH}$, TS3C$_{MV}$, $DD_{DTW}$-HC and KSC, respectively.

\begin{table*}[htp]
	\centering
	\caption{RI performance and computational time of the different algorithms for all the datasets. TS3C$_{CH}$: TS3C algorithm using CH as strategy, TS3C$_{MV}$: TS3C algorithm using MV as strategy, $DD_{DTW}$-HC: hierarchical clustering using $DD_{DTW}$ distance, KSC: K-Spectral Centroid clustering algorithm. The best results for each dataset is highlighted in bold face. The parameter $SEP_{\max}$ for each dataset is shown as a subscript for the TS3C$_{CH}$ and TS3C$_{MV}$ methods. (1/2)}
	\label{table:resultsMV}
	\resizebox{\textwidth}{!}{
		\begin{tabular}{ccccc||cccc}
			\hline\hline
			&\multicolumn{4}{c||}{Rand Index}& \multicolumn{4}{c}{Time (seconds)} \\ \cline{2-9}
			Dataset&TS3C$_{CH}$&TS3C$_{MV}$&$DD_{DTW}$-HC&KSC&TS3C$_{CH}$&TS3C$_{MV}$&$DD_{DTW}$-HC&KSC \\ \hline
			50W & $\mathbf{0.935}_{20}$ & $\mathbf{0.935}_{10}$ & $\mathit{0.923}$ & $0.655$ & $\mathbf{1478.045}$ & $\mathit{1503.025}$ & $104284.860$ & $17667.431$\\
			ADI & $\mathit{0.924}_{40}$ & $0.916_{50}$ & $0.683$ & $\mathbf{0.948}$ & $\mathbf{886.631}$ & $\mathit{903.577}$ & $32220.837$ & $5374.062$\\
			ARR & $0.619_{50}$ & $\mathbf{0.632}_{40}$ & $0.349$ & $\mathit{0.628}$ & $\mathit{361.145}$ & $361.694$ & $4603.193$ & $\mathbf{83.950}$\\
			BEE & $\mathit{0.680}_{100}$ & $\mathit{0.680}_{100}$ & $0.582$ & $\mathbf{0.708}$ & $\mathit{188.140}$ & $188.475$ & $1247.072$ & $\mathbf{62.618}$\\
			BFL & $0.492_{60}$ & $0.492_{60}$ & $\mathbf{0.591}$ & $\mathit{0.495}$ & $\mathit{136.334}$ & $136.498$ & $642.260$ & $\mathbf{20.673}$\\
			BIR & $0.492_{100}$ & $0.492_{100}$ & $\mathit{0.499}$ & $\mathbf{0.541}$ & $\mathit{116.866}$ & $117.151$ & $688.234$ & $\mathbf{18.569}$\\
			CAR & $\mathit{0.653}_{50}$ & $\mathit{0.653}_{50}$ & $0.498$ & $\mathbf{0.680}$ & $\mathit{391.834}$ & $392.027$ & $8371.195$ & $\mathbf{142.094}$\\
			CBF & $\mathit{0.673}_{10}$ & $\mathit{0.673}_{10}$ & $\mathbf{0.776}$ & $0.561$ & $\mathit{975.780}$ & $978.230$ & $24829.780$ & $\mathbf{584.100}$\\
			CHL & $\mathit{0.491}_{20}$ & $0.473_{10}$ & $0.403$ & $\mathbf{0.527}$ & $\mathit{6555.744}$ & $6601.373$ & $763586.744$ & $\mathbf{2623.021}$\\
			CIN & $\mathit{0.641}_{90}$ & $\mathit{0.641}_{90}$ & $0.555$ & $\mathbf{0.693}$ & $\mathbf{11647.877}$ & $\mathit{11652.715}$ & $>763587$ & $27771.510$\\
			COF & $\mathit{0.507}_{80}$ & $\mathit{0.507}_{80}$ & $0.491$ & $\mathbf{0.747}$ & $\mathit{97.762}$ & $98.123$ & $385.247$ & $\mathbf{10.141}$\\
			COM & $\mathit{0.502}_{40}$ & $\mathbf{0.509}_{100}$ & $0.500$ & $0.500$ & $\mathit{2698.562}$ & $2699.753$ & $210113.411$ & $\mathbf{413.442}$\\
			CRX & $\mathbf{0.847}_{20}$ & $\mathbf{0.847}_{20}$ & $\mathit{0.777}$ & $0.406$ & $\mathbf{1708.057}$ & $\mathit{1715.303}$ & $82969.617$ & $4355.741$\\
			CRY & $\mathbf{0.840}_{10}$ & $\mathbf{0.840}_{10}$ & $\mathit{0.688}$ & $0.529$ & $\mathbf{1751.838}$ & $\mathit{1759.408}$ & $83709.689$ & $4487.680$\\
			CRZ & $\mathit{0.835}_{20}$ & $\mathbf{0.845}_{100}$ & $0.710$ & $0.413$ & $\mathbf{1679.883}$ & $\mathit{1686.537}$ & $78563.430$ & $3806.989$\\
			DIA & $\mathit{0.720}_{40}$ & $\mathit{0.720}_{40}$ & $0.296$ & $\mathbf{0.958}$ & $\mathit{622.977}$ & $624.139$ & $20433.348$ & $\mathbf{302.917}$\\
			DPA & $0.597_{30}$ & $0.597_{30}$ & $\mathit{0.709}$ & $\mathbf{0.721}$ & $\mathit{312.147}$ & $313.675$ & $3091.080$ & $\mathbf{82.608}$\\
			DPC & $\mathit{0.506}_{20}$ & $\mathit{0.506}_{20}$ & $\mathbf{0.527}$ & $0.499$ & $\mathit{483.987}$ & $486.701$ & $8535.584$ & $\mathbf{45.135}$\\
			DPT & $\mathit{0.677}_{70}$ & $0.659_{90}$ & $\mathbf{0.862}$ & $0.660$ & $\mathit{303.372}$ & $305.925$ & $3231.426$ & $\mathbf{193.843}$\\
			EAR & $0.530_{10}$ & $0.529_{30}$ & $\mathit{0.541}$ & $\mathbf{0.615}$ & $\mathit{2246.077}$ & $2247.397$ & $93261.790$ & $\mathbf{307.636}$\\
			EC2 & $0.498_{70}$ & $0.498_{70}$ & $\mathit{0.537}$ & $\mathbf{0.612}$ & $\mathit{121.680}$ & $122.334$ & $650.748$ & $\mathbf{18.141}$\\
			EC5 & $\mathit{0.637}_{90}$ & $0.600_{50}$ & $\mathbf{0.891}$ & $0.591$ & $\mathbf{5019.159}$ & $\mathit{5088.073}$ & $736064.909$ & $16214.154$\\
			EFC & $\mathit{0.501}_{50}$ & $\mathit{0.501}_{50}$ & $0.499$ & $\mathbf{0.809}$ & $\mathit{942.266}$ & $944.596$ & $19578.035$ & $\mathbf{135.891}$\\
			ELE & $\mathbf{0.716}_{100}$ & $\mathbf{0.716}_{100}$ & $\mathit{0.441}$ & $0.362$ & $\mathit{11974.452}$ & $12619.793$ & $>763587$ & $\mathbf{2944.211}$\\
			FAA & $\mathbf{0.851}_{10}$ & $\mathbf{0.851}_{10}$ & $\mathit{0.604}$ & $0.302$ & $\mathbf{1875.987}$ & $\mathit{1892.960}$ & $>763587$ & $5236.007$\\
			FAF & $\mathbf{0.572}_{20}$ & $\mathbf{0.572}_{20}$ & $\mathit{0.550}$ & $0.384$ & $\mathit{269.547}$ & $269.889$ & $2448.621$ & $\mathbf{90.123}$\\
			FIS & $\mathit{0.729}_{80}$ & $0.639_{40}$ & $0.181$ & $\mathbf{0.792}$ & $\mathbf{930.099}$ & $\mathit{931.535}$ & $42706.796$ & $1264.597$\\
			FOA & $\mathit{0.523}_{80}$ & $0.514_{100}$ & $\mathbf{0.535}$ & $0.504$ & $\mathbf{16264.904}$ & $\mathit{16315.886}$ & $>763587$ & $36822.712$\\
			FOB & $\mathbf{0.500}_{20}$ & $\mathbf{0.500}_{20}$ & $\mathbf{0.500}$ & $\mathbf{0.500}$ & $\mathbf{12186.926}$ & $\mathit{12221.526}$ & $>763587$ & $28636.890$\\
			GUN & $\mathbf{0.540}_{70}$ & $\mathbf{0.540}_{70}$ & $0.498$ & $\mathit{0.507}$ & $\mathit{133.758}$ & $133.962$ & $2009.401$ & $\mathbf{9.739}$\\
			HAM & $\mathit{0.517}_{90}$ & $\mathit{0.517}_{90}$ & $0.498$ & $\mathbf{0.527}$ & $\mathit{556.913}$ & $557.675$ & $14038.784$ & $\mathbf{152.476}$\\
			HAN & $0.501_{50}$ & $\mathit{0.553}_{30}$ & $0.548$ & $\mathbf{0.685}$ & $\mathbf{20404.031}$ & $\mathit{20408.041}$ & $>763587$ & $34051.703$\\
			HAP & $\mathit{0.601}_{20}$ & $\mathit{0.601}_{20}$ & $0.389$ & $\mathbf{0.690}$ & $\mathbf{2302.345}$ & $\mathit{2303.288}$ & $>763587$ & $3881.453$\\
			HER & $0.498_{70}$ & $0.498_{70}$ & $\mathbf{0.514}$ & $\mathit{0.502}$ & $\mathit{273.273}$ & $273.411$ & $7475.375$ & $\mathbf{38.180}$\\
			INL & $\mathit{0.711}_{90}$ & $\mathit{0.711}_{90}$ & $0.535$ & $\mathbf{0.739}$ & $\mathbf{5598.252}$ & $\mathit{5600.156}$ & $>763587$ & $11234.116$\\
			INS & $\mathbf{0.813}_{10}$ & $\mathbf{0.813}_{10}$ & $0.548$ & $\mathit{0.690}$ & $\mathbf{3661.013}$ & $\mathit{3685.342}$ & $474905.381$ & $10472.539$\\
			ITA & $0.500_{70}$ & $0.500_{70}$ & $\mathit{0.512}$ & $\mathbf{0.639}$ & $\mathit{242.212}$ & $245.741$ & $1233.470$ & $\mathbf{19.684}$\\
			LAR & $\mathbf{0.549}_{30}$ & $\mathbf{0.549}_{30}$ & $0.343$ & $\mathit{0.412}$ & $\mathit{3906.045}$ & $3908.310$ & $369379.309$ & $\mathbf{710.732}$\\
			LI2 & $0.500_{10}$ & $\mathbf{0.538}_{20}$ & $0.500$ & $\mathit{0.503}$ & $\mathit{543.498}$ & $543.752$ & $9477.971$ & $\mathbf{127.702}$\\
			LI7 & $\mathbf{0.751}_{10}$ & $\mathbf{0.751}_{10}$ & $\mathit{0.604}$ & $0.590$ & $\mathit{323.328}$ & $323.818$ & $3124.774$ & $\mathbf{179.850}$\\
			MAL & $0.802_{70}$ & $0.802_{70}$ & $\mathbf{0.926}$ & $\mathit{0.916}$ & $\mathbf{14509.726}$ & $\mathit{14529.593}$ & $>763587$ & $18388.235$\\
			MEA & $0.711_{70}$ & $0.403_{50}$ & $\mathbf{0.770}$ & $\mathit{0.760}$ & $\mathit{279.293}$ & $279.455$ & $4717.715$ & $\mathbf{85.614}$\\
			\hline\hline
			\multicolumn{9}{l}{The best result is highlighted in bold face, while the second one is shown in italics}
		\end{tabular}
	}
\end{table*}

\begin{table*}[htp]
	\centering
	\setcounter{table}{1}
	\caption{RI performance and computational time of the different algorithms for all the datasets. TS3C$_{CH}$: TS3C algorithm using CH as strategy, TS3C$_{MV}$: TS3C algorithm using MV as strategy, $DD_{DTW}$-HC: hierarchical clustering using $DD_{DTW}$ distance, KSC: K-Spectral Centroid clustering algorithm. The best results for each dataset is highlighted in bold face. The parameter $SEP_{\max}$ for each dataset is shown as a subscript for the TS3C$_{CH}$ and TS3C$_{MV}$ methods. (2/2)}
	\label{table:resultsMV2}
	\resizebox{\textwidth}{!}{
		\begin{tabular}{ccccc||cccc}
			\hline\hline
			&\multicolumn{4}{c||}{Rand Index}& \multicolumn{4}{c}{Time (seconds)} \\ \cline{2-4} \cline{5-9}
			Dataset&TS3C$_{CH}$&TS3C$_{MV}$&$DD_{DTW}$-HC&KSC&TS3C$_{CH}$&TS3C$_{MV}$&$DD_{DTW}$-HC&KSC \\ \hline
			MED & $\mathit{0.649}_{80}$ & $\mathbf{0.652}_{30}$ & $0.636$ & $0.471$ & $\mathbf{646.388}$ & $\mathit{651.434}$ & $19953.705$ & $1062.241$\\
			MPA & $0.563_{90}$ & $0.563_{90}$ & $\mathit{0.729}$ & $\mathbf{0.731}$ & $\mathit{249.724}$ & $250.551$ & $4085.933$ & $\mathbf{70.214}$\\
			MPC & $\mathbf{0.505}_{100}$ & $\mathbf{0.505}_{100}$ & $\mathit{0.500}$ & $\mathit{0.500}$ & $\mathit{394.865}$ & $396.213$ & $9306.403$ & $\mathbf{64.696}$\\
			MPT & $0.741_{90}$ & $\mathbf{0.820}_{30}$ & $0.802$ & $\mathit{0.805}$ & $\mathit{286.181}$ & $288.578$ & $3318.997$ & $\mathbf{204.006}$\\
			MOT & $0.500_{90}$ & $0.500_{90}$ & $\mathit{0.503}$ & $\mathbf{0.580}$ & $\mathit{716.729}$ & $720.674$ & $20455.903$ & $\mathbf{496.467}$\\
			NO1 & $0.940_{20}$ & $\mathit{0.951}_{10}$ & $0.704$ & $\mathbf{0.953}$ & $\mathbf{17115.433}$ & $\mathit{17225.823}$ & $>763587$ & $218649.675$\\
			NO2 & $\mathit{0.952}_{60}$ & $0.950_{10}$ & $0.851$ & $\mathbf{0.965}$ & $\mathbf{13593.782}$ & $\mathit{13676.459}$ & $>763587$ & $208415.905$\\
			OLI & $\mathit{0.774}_{30}$ & $\mathit{0.774}_{30}$ & $0.758$ & $\mathbf{0.851}$ & $\mathit{138.726}$ & $138.842$ & $2443.925$ & $\mathbf{66.271}$\\
			OSU & $\mathbf{0.730}_{10}$ & $\mathbf{0.730}_{10}$ & $\mathit{0.619}$ & $0.290$ & $\mathit{1097.406}$ & $1099.504$ & $60765.041$ & $\mathbf{670.476}$\\
			PHA & $\mathit{0.514}_{70}$ & $\mathit{0.514}_{70}$ & $\mathbf{0.539}$ & $0.506$ & $\mathit{1499.161}$ & $1515.496$ & $77557.691$ & $\mathbf{200.132}$\\
			PHO & $\mathbf{0.928}_{60}$ & $\mathit{0.927}_{100}$ & $0.453$ & $0.508$ & $\mathbf{15477.269}$ & $\mathit{15524.564}$ & $>763587$ & $364702.702$\\
			PLA & $0.828_{10}$ & $0.795_{60}$ & $\mathbf{1.000}$ & $\mathit{0.917}$ & $\mathit{148.663}$ & $149.215$ & $2046.920$ & $\mathbf{52.744}$\\
			PPA & $0.757_{60}$ & $0.757_{60}$ & $\mathbf{0.779}$ & $\mathit{0.762}$ & $\mathit{234.914}$ & $236.040$ & $4908.688$ & $\mathbf{82.133}$\\
			PPC & $\mathbf{0.563}_{70}$ & $\mathit{0.559}_{50}$ & $0.535$ & $0.533$ & $\mathit{418.170}$ & $420.871$ & $8195.188$ & $\mathbf{31.829}$\\
			PPT & $0.783_{80}$ & $0.783_{80}$ & $\mathbf{0.880}$ & $\mathit{0.805}$ & $\mathit{319.035}$ & $320.944$ & $3850.762$ & $\mathbf{220.865}$\\
			REF & $\mathbf{0.564}_{40}$ & $\mathit{0.537}_{20}$ & $0.352$ & $0.393$ & $\mathit{3969.303}$ & $3971.592$ & $422843.191$ & $\mathbf{900.040}$\\
			SCR & $\mathbf{0.533}_{100}$ & $\mathbf{0.533}_{100}$ & $0.346$ & $\mathit{0.452}$ & $\mathit{4022.248}$ & $4024.291$ & $355109.311$ & $\mathbf{1357.686}$\\
			SHS & $\mathbf{0.990}_{100}$ & $\mathbf{0.990}_{100}$ & $\mathit{0.498}$ & $\mathit{0.498}$ & $\mathit{1005.236}$ & $1005.601$ & $13878.526$ & $\mathbf{111.026}$\\
			SHA & $\mathbf{0.969}_{90}$ & $\mathbf{0.969}_{90}$ & $\mathit{0.838}$ & $0.628$ & $\mathbf{3737.862}$ & $\mathit{3776.948}$ & $546585.181$ & $51106.723$\\
			SMA & $\mathbf{0.592}_{10}$ & $\mathbf{0.592}_{10}$ & $0.339$ & $\mathit{0.537}$ & $\mathit{4047.678}$ & $4050.231$ & $379869.334$ & $\mathbf{1487.027}$\\
			SO1 & $0.514_{90}$ & $\mathit{0.523}_{70}$ & $0.499$ & $\mathbf{0.751}$ & $\mathit{348.820}$ & $350.155$ & $2769.672$ & $\mathbf{73.918}$\\
			SO2 & $\mathit{0.599}_{100}$ & $0.530_{10}$ & $0.534$ & $\mathbf{0.657}$ & $\mathit{607.075}$ & $610.305$ & $9039.759$ & $\mathbf{149.099}$\\
			STA & $0.595_{90}$ & $0.595_{90}$ & $\mathbf{0.833}$ & $\mathit{0.769}$ & $\mathit{38682.681}$ & $38815.934$ & $>763587$ & $\mathbf{32802.621}$\\
			STR & $0.502_{50}$ & $\mathbf{0.519}_{40}$ & $\mathit{0.504}$ & $\mathit{0.504}$ & $\mathit{1330.742}$ & $1333.893$ & $92051.786$ & $\mathbf{365.554}$\\
			SWE & $\mathit{0.875}_{60}$ & $\mathbf{0.880}_{50}$ & $0.348$ & $0.630$ & $\mathbf{984.759}$ & $\mathit{995.990}$ & $36092.199$ & $1619.722$\\
			SYM & $\mathit{0.810}_{80}$ & $\mathit{0.810}_{80}$ & $\mathbf{0.886}$ & $0.604$ & $\mathbf{2773.593}$ & $\mathit{2778.688}$ & $248453.812$ & $2867.344$\\
			SYN & $\mathit{0.782}_{100}$ & $\mathit{0.782}_{100}$ & $\mathbf{0.875}$ & $0.384$ & $\mathit{361.445}$ & $364.580$ & $2112.292$ & $\mathbf{235.131}$\\
			TO1 & $\mathit{0.514}_{70}$ & $\mathit{0.514}_{70}$ & $0.505$ & $\mathbf{0.533}$ & $\mathit{526.967}$ & $527.858$ & $8315.421$ & $\mathbf{121.068}$\\
			TO2 & $0.497_{40}$ & $0.497_{40}$ & $\mathbf{0.665}$ & $\mathit{0.532}$ & $\mathit{338.448}$ & $339.017$ & $5257.642$ & $\mathbf{79.495}$\\
			TRA & $\mathit{0.843}_{40}$ & $\mathit{0.843}_{40}$ & $\mathbf{0.874}$ & $0.721$ & $\mathit{324.761}$ & $325.294$ & $4986.481$ & $\mathbf{113.633}$\\
			TWP & $\mathit{0.644}_{10}$ & $\mathit{0.644}_{10}$ & $\mathbf{0.848}$ & $0.455$ & $\mathit{6464.721}$ & $6531.603$ & $581050.498$ & $\mathbf{4601.802}$\\
			TWE & $\mathbf{0.644}_{10}$ & $\mathbf{0.644}_{10}$ & $0.500$ & $\mathit{0.541}$ & $\mathit{702.708}$ & $706.575$ & $11902.970$ & $\mathbf{240.079}$\\
			UWZ & $\mathit{0.775}_{100}$ & $0.754_{10}$ & $\mathbf{0.800}$ & $0.506$ & $\mathbf{7470.248}$ & $\mathit{7523.152}$ & $>763587$ & $47423.308$\\
			UWY & $\mathit{0.781}_{90}$ & $0.759_{70}$ & $\mathbf{0.819}$ & $0.535$ & $\mathbf{8951.039}$ & $\mathit{9013.533}$ & $>763587$ & $48750.542$\\
			UWZ & $\mathbf{0.795}_{90}$ & $\mathbf{0.795}_{90}$ & $\mathit{0.735}$ & $0.541$ & $\mathbf{7264.171}$ & $\mathit{7314.869}$ & $>763587$ & $47553.727$\\
			UWA & $\mathbf{0.759}_{20}$ & $\mathbf{0.759}_{20}$ & $\mathit{0.590}$ & $0.447$ & $\mathbf{18754.477}$ & $\mathit{18799.317}$ & $>763587$ & $167079.000$\\
			WAF & $0.502_{70}$ & $\mathbf{0.655}_{10}$ & $0.534$ & $\mathit{0.591}$ & $\mathit{4087.145}$ & $4157.535$ & $>763587$ & $\mathbf{1681.639}$\\
			WIN & $\mathit{0.571}_{70}$ & $0.499_{90}$ & $0.499$ & $\mathbf{0.591}$ & $\mathit{104.940}$ & $105.030$ & $1273.336$ & $\mathbf{30.807}$\\
			WOS & $\mathit{0.870}_{20}$ & $\mathit{0.870}_{20}$ & $\mathbf{0.872}$ & $0.496$ & $\mathbf{1463.361}$ & $\mathit{1476.260}$ & $98232.829$ & $8212.427$\\
			WOR & $\mathit{0.597}_{20}$ & $0.582_{60}$ & $\mathbf{0.616}$ & $0.525$ & $\mathit{1604.660}$ & $1605.702$ & $82527.057$ & $\mathbf{1084.796}$\\
			WOT & $\mathbf{0.506}_{20}$ & $\mathbf{0.506}_{20}$ & $\mathit{0.503}$ & $0.499$ & $\mathit{1620.145}$ & $1620.900$ & $77497.308$ & $\mathbf{676.226}$\\
			YOG & $\mathbf{0.514}_{90}$ & $0.500_{50}$ & $\mathit{0.504}$ & $0.500$ & $\mathit{7959.012}$ & $7983.016$ & $>763587$ & $\mathbf{4847.798}$\\
			\hline\hline
			\multicolumn{9}{l}{The best result is highlighted in bold face, while the second one is shown in italics}
		\end{tabular}
	}
\end{table*}

\subsection{Statistical analysis}
Based on the previous results, we consider all datasets to apply a set of non-parametric statistical tests in order to determine whether the differences found are obtained by chance. Given that the mean values across all datasets do not follow a normal distribution, we run the Wilcoxon signed-rank test, which is a nonparametric test that can be used to determine whether two dependent samples were selected from populations having the same distribution \cite{Wilcoxon1945,Hochberg1987}. This design for the statistical tests makes possible the comparison of the deterministic methods (TS3C$_{CH}$, TS3C$_{MV}$ and $DD_{DTW}$-HC) with the stochastic method (KSC, for which the average RI from the $30$ runs is used).

Results of the tests made using average RI are shown in \tablename{ \ref{table:testsRI}}. As can be observed, the differences are statistically significant for $\alpha=0.05$ between TS3C$_{CH}$ and $DD_{DTW}$-HC, and between TS3C$_{MV}$ and $DD_{DTW}$-HC. Also, if we consider $\alpha=0.10$, the methodology TS3C$_{MV}$ is statistically better than KSC. Consequently, these results show that the proposed methodology obtains more robust results than these state-of-the-art alternatives.

\begin{table*}[htp]
	\centering
	\caption{Wilcoxon tests for the comparison of the different  algorithms: adjusted $p$-values, using average RI as the test variable}
	\label{table:testsRI}
	\resizebox{0.87\textwidth}{!}{
		\begin{tabular}{ccccccc}
			\hline\hline
			\multirow{2}{*}{ } & TS3C$_{MV}$ vs & $DD_{DTW}$-HC vs & KSC vs & $DD_{DTW}$-HC vs & KSC vs & KSC vs \\
			& TS3C$_{CH}$ & TS3C$_{CH}$ & TS3C$_{CH}$ & TS3C$_{MV}$ & TS3C$_{MV}$ & $DD_{DTW}$-HC\\ \hline
			z-score & $-1.225$ & $-2.395$  & $-1.787$ & $-2.476$ & $-1.900$ & $-0.281$ \\
			p-value & $0.226$  &  $0.016$(*)  &  $0.074$(+) &  $0.013$(*) &  $0.057$(+) &  $0.781$ \\
			\hline\hline
			\multicolumn{7}{l}{* : Significant differences were found for $\alpha=$0.05.}\\
			\multicolumn{7}{l}{+: Significant differences were found for $\alpha=$0.10.}
		\end{tabular}
	}
\end{table*}

On the other side, results of the tests made using average computational time are shown in \tablename{ \ref{table:teststime}}. In this case, considering $\alpha=0.05$, there are statistically significant differences between: TS3C$_{MV}$ and TS3C$_{CH}$, TS3C$_{CH}$ and $DD_{DTW}$-HC, TS3C$_{MV}$ and $DD_{DTW}$-HC and, KSC and $DD_{DTW}$-HC. This means that both TS3C methods are more efficient than $DD_{DTW}$-HC, and that there are no significant differences when comparing them to KSC.

\begin{table*}[htp]
	\centering
	\caption{Wilcoxon tests for the comparison of the different  algorithms: adjusted $p$-values, using average time as the test variable}
	\label{table:teststime}
	\resizebox{0.87\textwidth}{!}{
		\begin{tabular}{ccccccc}
			\hline\hline
			\multirow{2}{*}{ } & TS3C$_{MV}$ vs & $DD_{DTW}$-HC vs & KSC vs & $DD_{DTW}$-HC vs & KSC vs & KSC vs \\
			& TS3C$_{CH}$ & TS3C$_{CH}$ & TS3C$_{CH}$ & TS3C$_{MV}$ & TS3C$_{MV}$ & $DD_{DTW}$-HC\\ \hline
			z-score & $-7.961$ & $-7.961$  & $-0.107$ & $-7.961$ & $-0.103$ & $-7.961$ \\
			p-value & $0.000$(*)  &  $0.000$(*)  &  $0.917$ &  $0.000$(*) &  $0.920$ &  $0.000$(*) \\ 
			\hline\hline
			\multicolumn{7}{l}{* : Significant differences were found for $\alpha=$0.05.}
		\end{tabular}
	}
\end{table*}

\section{Conclusions} \label{sec:conclusions}

In this paper, we have presented and tested a novel time series clustering approach, for the purpose of exploiting the similarities that can be found in subsequences of the time series analysed. The method is a two-stage statistical segmentation-clustering time series procedure, TS3C, which is based on: (1) a least squares polynomial segmentation procedure, using the growing window method, (2) an extraction of features of each segment (polynomial trend coefficients, variance, skewness and autocorrelation coefficient), (3) a clustering of these features using a hierarchical clustering, (4) a representation of each cluster by its centroid, the segment with higher variance, the difference in MSE, and the number of segments, (5) a mapping of the time series using the information of its clusters, and (6) a final clustering stage using the mapped dataset as input. Internal performance measures are used to adjust the only parameter value.

The proposed TS3C method is compared against two state-of-the-art methods: hierarchical clustering using the $DD_{DTW}$ distance measure ($DD_{DTW}$-HC) and the K-Spectral Centroid clustering algorithm ($KSC$). Our method outperforms both methods using two different approaches for deciding the values of the parameters. Although the segmentation process and the first hierarchical clustering involves a considerable computational load, the global cost is acceptable, given that the final clustering does not depend on the size of the original time series. In addition, a Wilcoxon signed-rank test statistical test is used to evaluate whether that the methodology is statistically more accurate and/or more efficient than the state-of-the-art algorithms.

A future line of research corresponds to the use of different approximation methods and segmentation techniques, with the purpose of reducing the computational cost of the first stage. Another direction can be the application of this methodology as previous step for prediction tasks (ordinal or nominal classification).

\section*{Acknowledgment}
This work has been subsidised by the by the TIN2014-54583-C2-1-R and the TIN2015-70308-REDT projects of the Spanish Ministry of Economy and Competitiveness (MINECO) and FEDER funds (EU). David Guijo-Rubio's research has been subsidised by the project PI15/01570 of the Fundación de Investigación Biomédica de Córdoba (FIBICO) and by the FPU Predoctoral Program (Spanish Ministry of Education and Science), grant reference FPU16/02128. Antonio M. Durán-Rosal's research is supported by the FPU Predoctoral Program (Spanish Ministry of Education and Science), grant reference FPU14/03039.

\ifCLASSOPTIONcaptionsoff
  \newpage
\fi


%

\bibliographystyle{IEEEtran}
\bibliography{bibliography}


%
%
%

%

\begin{IEEEbiography}[{\includegraphics[width=1in,height=1.25in,clip,keepaspectratio]{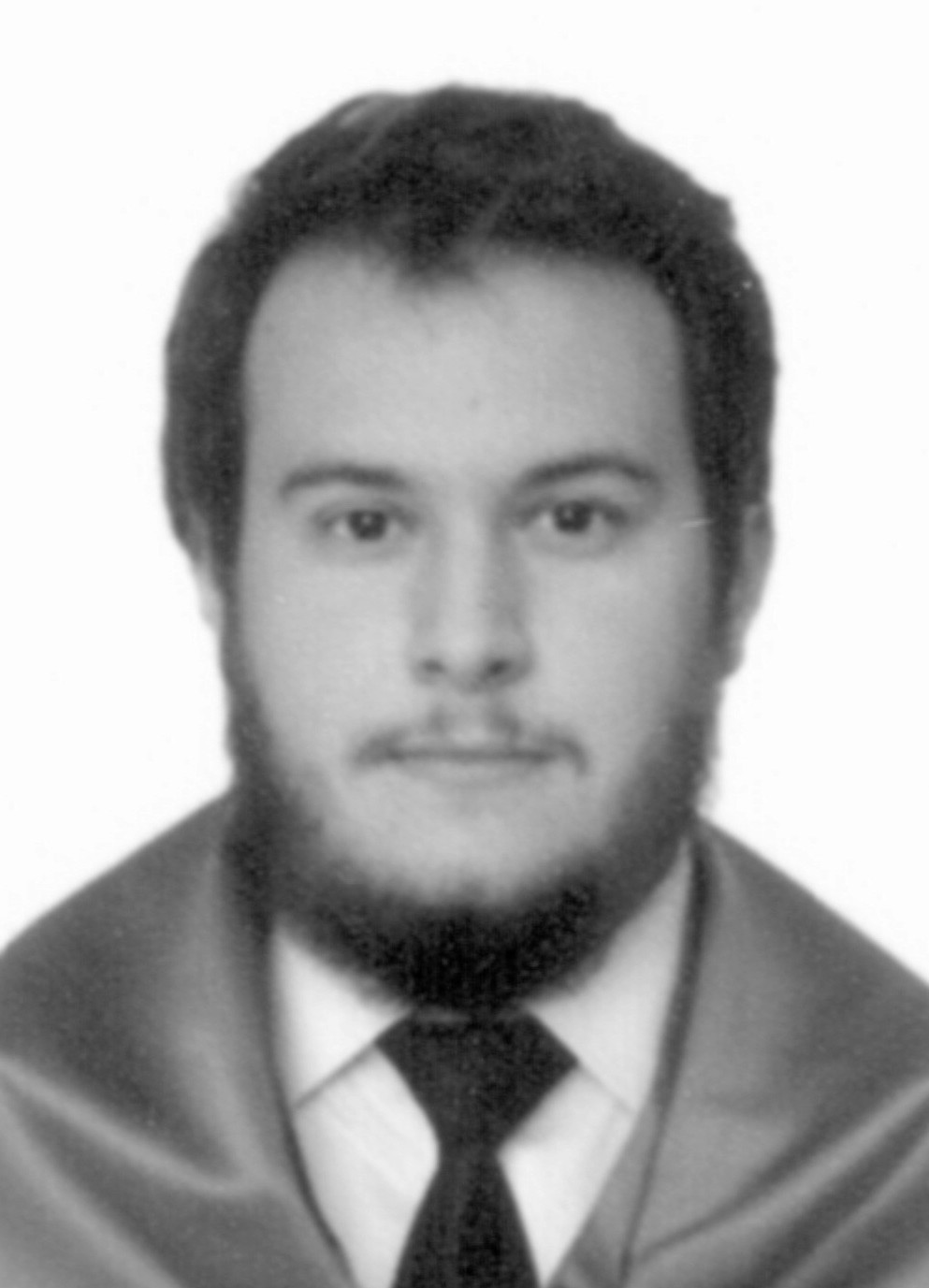}}]{David Guijo-Rubio}
	received the BS degree in computer science from the University of C{\'o}rdoba, Spain, in 2016, and the MSc degree in artificial intelligence from the International University Menendez Pelayo, Santander, in 2017. He is currently pursuing the Ph.D. degree in Computer Science and Artificial Intelligence in the Department of Computer Science and Numerical Analysis. His current research interests include a wide range of topics concerning machine learning and time series data mining.
\end{IEEEbiography}

\begin{IEEEbiography}[{\includegraphics[width=1in,height=1.25in,clip,keepaspectratio]{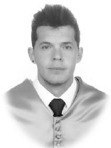}}]{Antonio Manuel Dur{\'a}n-Rosal}
	received the B.S. degree in Computer Science in 2014 and the M.Sc. degree in Computer Science in 2016 from the University of Córdoba, Spain, where he is currently pursuing the Ph.D. degree in Computer Science and Artificial Intelligence in the Department of Computer Science and Numerical Analysis. His current interests include a wide range of topics concerning machine learning, pattern recognition and time series analysis.
\end{IEEEbiography}

\begin{IEEEbiography}[{\includegraphics[width=1in,height=1.25in,clip,keepaspectratio]{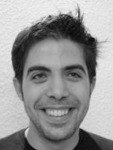}}]{Pedro Antonio Guti{\'e}rrez}
	received the BS degree in computer science from the University of Sevilla, Spain, in 2006, and the PhD degree in computer science and artificial intelligence from the University of Granada, Spain, in 2009. He is currently an assistant professor in the Department of Computer Science and Numerical Analysis, University of C{\'o}rdoba, Spain. His current research interests include pattern recognition, evolutionary computation, and their applications. He is a senior member of the IEEE.
\end{IEEEbiography}

\begin{IEEEbiography}[{\includegraphics[width=1in,height=1.25in,clip,keepaspectratio]{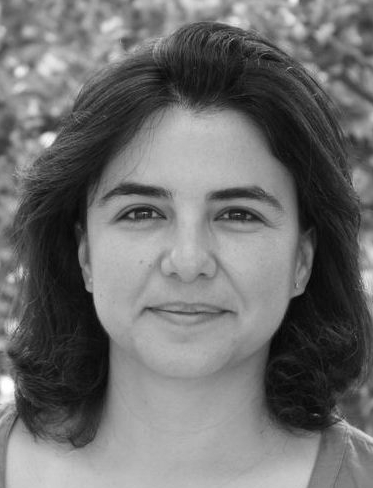}}]{Alicia Troncoso}
	received the Ph.D. degree in Computer Science from the University of Seville, Spain, in 2005. She was an Assistant Professor in the Department of Computer Science at the University of Seville from 2002 to 2005. She has been with the Department of Computer Science at the Pablo de Olavide University since 2005, where she is currently an Associate Professor. Her primary areas of interest are time series forecasting, data mining and big data. 
\end{IEEEbiography}

\begin{IEEEbiography}[{\includegraphics[width=1in,height=1.25in,clip,keepaspectratio]{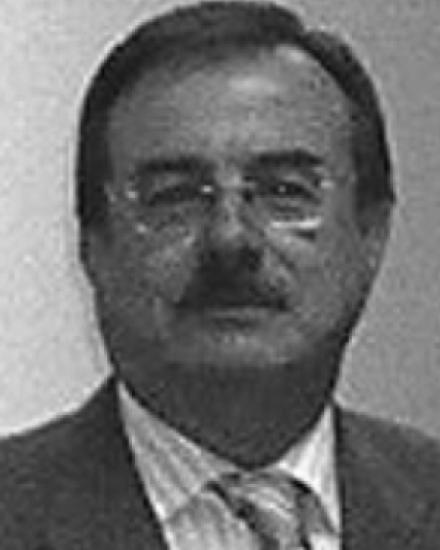}}]{C{\'e}sar Herv{\'a}s-Mart{\'i}nez}
	received the BS degree in statistics and operations research from the ``Universidad Complutense'', Madrid, Spain, in 1978, and the PhD degree in mathematics from the University of Seville, Spain, in 1986. He is currently a professor of computer science and artificial intelligence  in  the  Department  of  Computer Science and Numerical Analysis, University of C{\'o}rdoba, and an associate professor in the Department of Mathematics and Quantitative Methods, Loyola Andaluc{\'i}a University. His current research interests include neural networks, evolutionary computation, and the modeling of natural systems. He is a senior member of the IEEE.
\end{IEEEbiography}




\end{document}